\documentclass[letterpaper]{article} % DO NOT CHANGE THIS
\usepackage[preprint]{aaai2027}  % DO NOT CHANGE THIS
\usepackage[hyphens]{url}  % DO NOT CHANGE THIS
\usepackage{graphicx} % DO NOT CHANGE THIS
\usepackage{natbib}  % DO NOT CHANGE THIS AND DO NOT ADD ANY OPTIONS TO IT
\usepackage{caption} % DO NOT CHANGE THIS AND DO NOT ADD ANY OPTIONS TO IT
\usepackage{algorithm}
\usepackage{algorithmic}
\usepackage{booktabs}
\usepackage{multirow}
\usepackage{amssymb}
\usepackage{amsmath}
\usepackage{array}
\usepackage{tabularx}
\usepackage{xcolor}

\usepackage{newfloat}
\usepackage{listings}

\DeclareCaptionStyle{ruled}{
  labelfont=normalfont,
  labelsep=colon,
  strut=off
}

\lstdefinestyle{prompt}{
  basicstyle=\footnotesize\rmfamily,
  numbers=none,
  breaklines=true,
  breakatwhitespace=true,
  columns=fullflexible,
  keepspaces=true,
  showstringspaces=false,
  frame=leftline,
  framerule=0.4pt,
  rulecolor=\color{black!30},
  backgroundcolor=\color{black!2},
  framesep=4pt,
  framexleftmargin=4pt,
  xleftmargin=5pt,
  xrightmargin=1pt,
  aboveskip=2pt,
  belowskip=6pt
}

\newcommand{\prompttitle}[1]{%
  \par\smallskip
  \noindent\textbf{#1}\par
  \nopagebreak[4]
}

\floatstyle{ruled}
\newfloat{listing}{tb}{lst}{}
\floatname{listing}{Listing}

\title{CDEG: Learning Decision-Critical Evidence for Long-Horizon Diagnostic Agents}

\author{
Xiwei Dai\equalcontrib,
Zijie Meng\equalcontrib,
Zhiting Fan,
Yixuan Tang,
Guanyu Jiang,
Ziru Niu,
Zuozhu Liu\corresponding
}

\affiliations{
Zhejiang University\\
Hangzhou, Zhejiang, China
}

\begin{document}

\maketitle

\begin{abstract}
Unlike static medical question answering, long-horizon diagnosis captures the sequential nature of clinical practice: evidence is progressively acquired, integrated, and evaluated over multiple rounds of interaction before reaching a final diagnosis. However, existing doctor agents often fail when critical evidence is either not acquired or not adequately incorporated into diagnostic reasoning. Recent agentic approaches attempt to address these failures by reusing historical trajectories or distilled memories. But their diagnostic gains remain constrained because such experience may contain noisy or incidental information and is typically reused without validating which evidence actually drives diagnostic decisions. To address this limitation, we introduce \textbf{CDEG}, a graph-based framework that learns reusable decision-critical evidence from historical diagnostic trajectories. CDEG contrasts successful and failed trajectories from the same case to identify candidate evidence, validates their diagnostic impact through controlled counterfactual interventions, and organizes the resulting diagnosis--evidence--action relations into a structured graph. During inference, CDEG tracks the evolving patient evidence state to retrieve relevant diagnostic relations and selectively guide missing evidence acquisition or overlooked evidence reappraisal. Across in-domain and out-of-distribution benchmarks with multiple doctor agent backbones, CDEG consistently improves diagnostic performance, achieving up to an 11.5\% accuracy gain over vanilla agents. These results demonstrate that reliable long-horizon diagnosis requires moving beyond trajectory-level experience reuse toward evidence-level learning of the factors that truly shape clinical decisions.

\end{abstract}

\begin{figure}[t]
    \centering
    \includegraphics[width=\columnwidth]{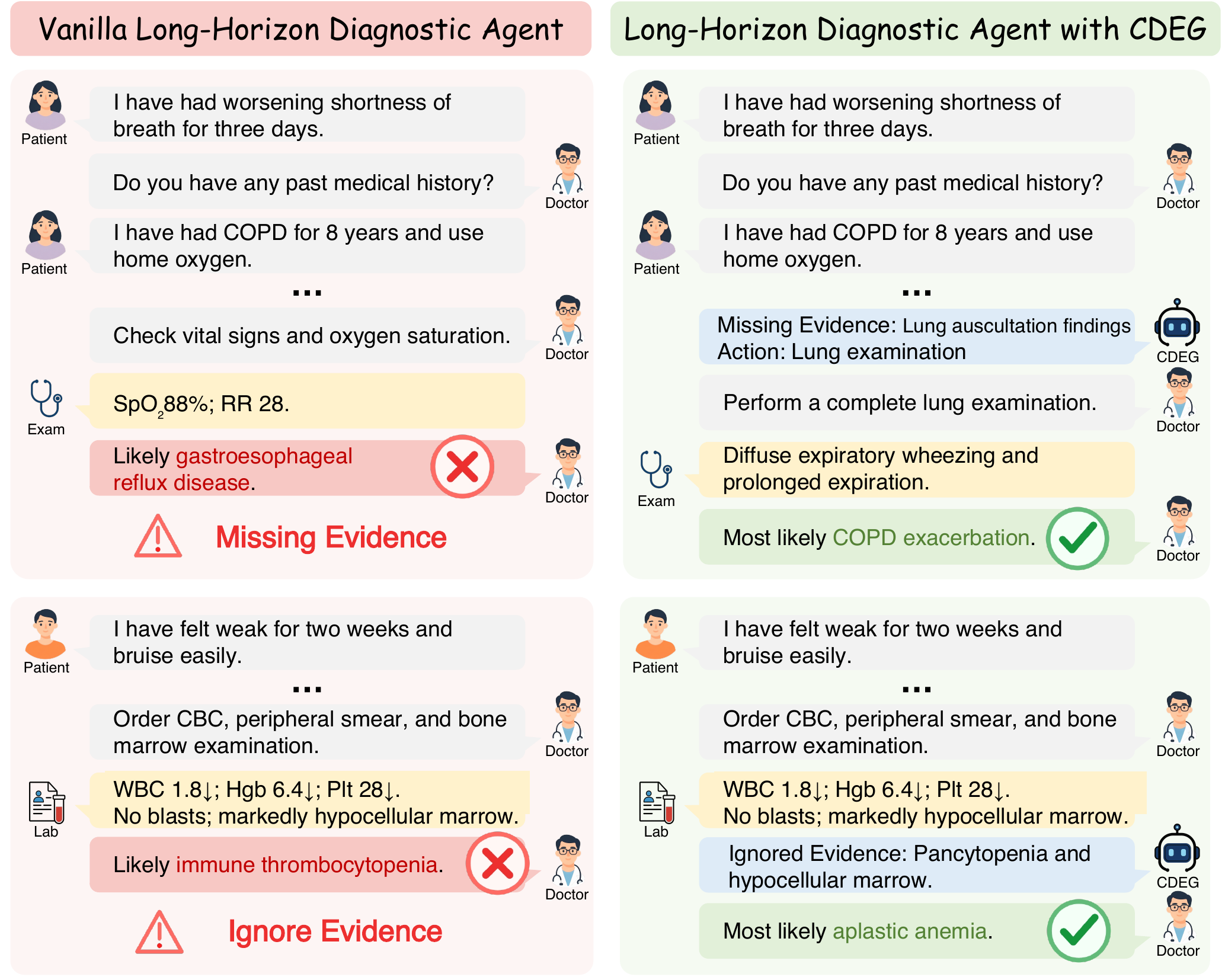}
    \caption{Long-horizon diagnosis with and without CDEG.
    The vanilla doctor agent may fail when decision-critical evidence is missing or ignored. With CDEG, the doctor agent is prompted to acquire missing evidence or reappraise observed evidence, enabling it to reach the correct diagnosis.}

    \label{fig:cdeg_overview}
\end{figure}
\section{Introduction}
\label{sec:introduction}

Recent advances in medical language models have shifted diagnostic agents beyond static medical question answering toward long-horizon interactive diagnosis \cite{agentclinic,medagentsim,mediq}. Under partial observability, a doctor agent must acquire evidence from the patient environment through diagnostic tools, integrate it across turns, and decide when it is sufficient for diagnosis \cite{nasem2015diagnosis}. Yet vanilla agents often fail to acquire relevant evidence or overlook observations already available \cite{evidenceelicitation,mediq,medaction,mintdiagnosis}, as illustrated in Figure~\ref{fig:cdeg_overview}. Therefore, a fundamental challenge in long-horizon diagnosis is how to identify and utilize the evidence that is truly critical for distinguishing diagnoses.

% Recent advances in medical language models have shifted diagnostic agents beyond static medical question answering toward long-horizon interactive diagnosis \cite{agentclinic,medagentsim,mediq}. In this setting, a doctor agent interacts with a patient environment and diagnostic tools over multiple turns under partial observability. Unlike conventional diagnosis with complete clinical information, the agent must progressively acquire evidence, integrate observations from previous interactions, and determine whether the collected evidence is sufficient for a final diagnosis \cite{nasem2015diagnosis}. However, existing doctor agents remain limited in effectively managing this process. They may fail to acquire clinically relevant evidence during interaction, or overlook already observed evidence when making diagnostic decisions \cite{evidenceelicitation,mediq,medaction,mintdiagnosis}. Therefore, a fundamental challenge in long-horizon diagnosis is how to identify and utilize the evidence that is truly critical for distinguishing competing diagnoses.

Historical diagnostic trajectories record evidence, actions, and final decisions, offering reusable experience for future interactions. Existing methods retrieve prior trajectories or distill them into reflections, strategies, and structured memories \cite{reflexion,expel,agentworkflowmemory,dxevolve,evomedagent,sea,gsem}. However, trajectories mix decision-relevant evidence with incidental observations, while outcomes do not reveal which evidence shaped the diagnosis. Recent methods attempt to improve experience reuse through trajectory refinement, applicability-aware retrieval, and memory management but still operate mainly at the trajectory or memory-unit level~\cite{reasoningbank,memorymanagement2026}. Thus, individual observations may be reused without establishing their diagnostic relevance.

% Historical diagnostic trajectories record how evidence was acquired, what actions were taken, and how diagnostic decisions were reached, providing valuable experience for future interactions. Existing approaches reuse such experience by retrieving previous trajectories or distilling them into reflections, strategies, and structured memories to guide subsequent decisions \cite{reflexion,expel,agentworkflowmemory,dxevolve,evomedagent,sea,gsem}. However, a diagnostic trajectory contains a mixture of relevant and incidental observations, and its final outcome cannot directly indicate which evidence contributed to the decision. Reusing trajectories without validating individual evidence may therefore introduce irrelevant or unreliable information into future reasoning. Recent methods attempt to improve experience reuse through trajectory refinement, applicability-aware retrieval, and utility-based memory management. However, these approaches still operate primarily at the trajectory or memory-unit level and lack the ability to determine the diagnostic relevance of individual clinical evidence. As a result, it remains unclear which evidence from prior trajectories is truly decision-critical and suitable for reuse.

Counterfactual reasoning offers a potential way to resolve this ambiguity \cite{richens2020counterfactual,nagesh2023counterfactual,xu2022counterfactual}. By changing one observation while keeping the remaining evidence unchanged, it tests whether diagnostic preference changes. We extend this principle to historical diagnostic trajectories and regard evidence as decision-critical only when a controlled intervention on its availability or use consistently changes diagnostic preference. This selects evidence for reuse by its effect on decisions, rather than its co-occurrence with a successful outcome.

% Counterfactual reasoning provides a potential evidence-level validation for resolving this ambiguity \cite{richens2020counterfactual,nagesh2023counterfactual,xu2022counterfactual}. By intervening on one clinical observation at a time while holding the remaining evidence fixed, it tests whether that observation changes the preference between competing diagnoses. We extend this principle to historical diagnostic trajectories and regard evidence as decision-critical only when a controlled intervention on its availability or use consistently changes diagnostic preference. This criterion selects evidence for reuse by its effect on diagnostic decisions, rather than its co-occurrence with a successful outcome.

Building on this criterion, we introduce \textbf{CDEG}, a graph-based framework for learning reusable decision-critical evidence from historical diagnostic trajectories. CDEG contrasts each failed trajectory with a similar successful trajectory from the same case, uses controlled counterfactual tests to determine whether candidate evidence supports or blocks the corresponding diagnostic revision, and organizes the resulting diagnosis--evidence--action relations into a structured graph. During inference, CDEG matches the patient's evolving evidence state to the graph to guide acquisition of missing evidence or reappraisal of overlooked observations. We evaluate CDEG on the in-domain (ID) MIMIC-IDx and out-of-distribution (OOD) AgentClinic benchmarks across four doctor-agent backbones~\cite{johnson2023mimic,mimicivnote,mimiccxr,agentclinic,jin2021medqa}. CDEG improves average accuracy by 7.63\% and 8.88\% on the two benchmarks, respectively, with gains of up to 11.5\%, consistently outperforming the vanilla agent and other experience-reuse methods. These results show that evidence-level learning provides a more reliable basis for experience reuse in long-horizon diagnosis. Our contributions are summarized as follows:

% Building on this criterion, we introduce \textbf{CDEG}, a graph-based framework for learning reusable decision-critical evidence from historical diagnostic trajectories. For each failed trajectory, CDEG identifies a similar successful trajectory from the same case and compares the evidence available in the two trajectories to find observations that were missing, overlooked, or potentially misleading. It then uses counterfactual validation to determine whether each candidate support or block the corresponding diagnostic revision under controlled interventions, and organizes the resulting diagnosis--evidence--action relations into a structured graph. During inference, CDEG matches the patient's evolving evidence state against the graph to guide further evidence acquisition or prompt reconsideration of existing observations. We evaluate CDEG on MIMIC-IDx and AgentClinic with four doctor agent backbones~\cite{johnson2023mimic,mimicivnote,mimiccxr,agentclinic,jin2021medqa}. CDEG improves average accuracy by 7.63\% and 8.88\% across backbones on the two benchmarks, respectively, with gains of up to 11.5\%, consistently outperforming the vanilla agent and other experience-reuse methods. These results show that identifying and validating individual evidence provides a more reliable basis for experience reuse in long-horizon diagnosis. Our contributions are summarized as follows:

\begin{itemize}

\item We formulate trajectory reuse as an evidence-level learning problem and introduce CDEG to extract decision-critical evidence from historical diagnostic trajectories.

\item We develop a counterfactual validation method that contrasts successful and failed trajectories from the same case and retains only evidence that demonstrably affects diagnostic preference under controlled interventions.

\item We construct a structured graph that uses the patient's evolving evidence state to guide acquisition of missing evidence or reappraisal of overlooked evidence.

\item We evaluate CDEG on ID and OOD benchmarks with various backbones, which consistently improves the vanilla agent and achieves the best diagnostic performance.

\end{itemize}

\section{Related Work}

\paragraph{Long-Horizon Interactive Diagnosis.}
Long-horizon interactive diagnosis requires a doctor agent to acquire and integrate clinical evidence under partial observability before providing a final diagnosis. Interactive environments and benchmarks instantiate this setting through multi-turn patient interaction and diagnostic tools~\cite{agentclinic,medagentsim,medagentbench,mediq}. Recent methods improve evidence acquisition and diagnostic reasoning through targeted questioning, test selection, and iterative diagnosis~\cite{diagagent,medexagent,medxagent,medaction,meddxagent,diallms,doccha}. Despite this progress, evidence acquisition remains challenging, and evidence already observed may still be ignored in the final diagnosis~\cite{evidenceelicitation,mintdiagnosis}.

\paragraph{Learning from Diagnostic Trajectories.}
A complementary line of works learns from prior trajectories to improve subsequent decisions. General agent methods retrieve trajectories as exemplars or distill them into verbal reflections, transferable lessons, reusable workflows, skill libraries, or reasoning memories~\cite{synapse,reflexion,expel,agentworkflowmemory,voyager,reasoningbank}. Medical agents similarly retrieve prior cases or distill diagnostic trajectories into strategies and structured memories~\cite{dxevolve,evomedagent,sea,gsem}. However, not all such experience is equally relevant or reliable for subsequent diagnosis, and its quality is not always explicitly assessed before reuse.

\paragraph{Counterfactual Reasoning in Medical Diagnosis.}
Counterfactual reasoning examines how diagnostic preference changes under controlled interventions on clinical evidence. Prior medical works use counterfactual inference and evidence perturbation for causal diagnosis, clinical prediction, and model explanation~\cite{richens2020counterfactual,nagesh2023counterfactual,xu2022counterfactual}. Recent medical-LLM studies further modify symptoms, laboratory evidence, or diagnostic contexts to evaluate diagnostic anchoring and reasoning over competing diagnoses~\cite{clinicalcausaleval,medeinst,medcounterfact,counterfactualmultiagent,medievalbenchmark}. These works use counterfactual reasoning mainly for case-level diagnosis, explanation, or evaluation. Our work instead applies counterfactual validation during graph construction to assess candidate evidence mined from successful and failed trajectories before incorporating them into CDEG.

\begin{figure*}[t]
    \centering
    \includegraphics[width=\textwidth]{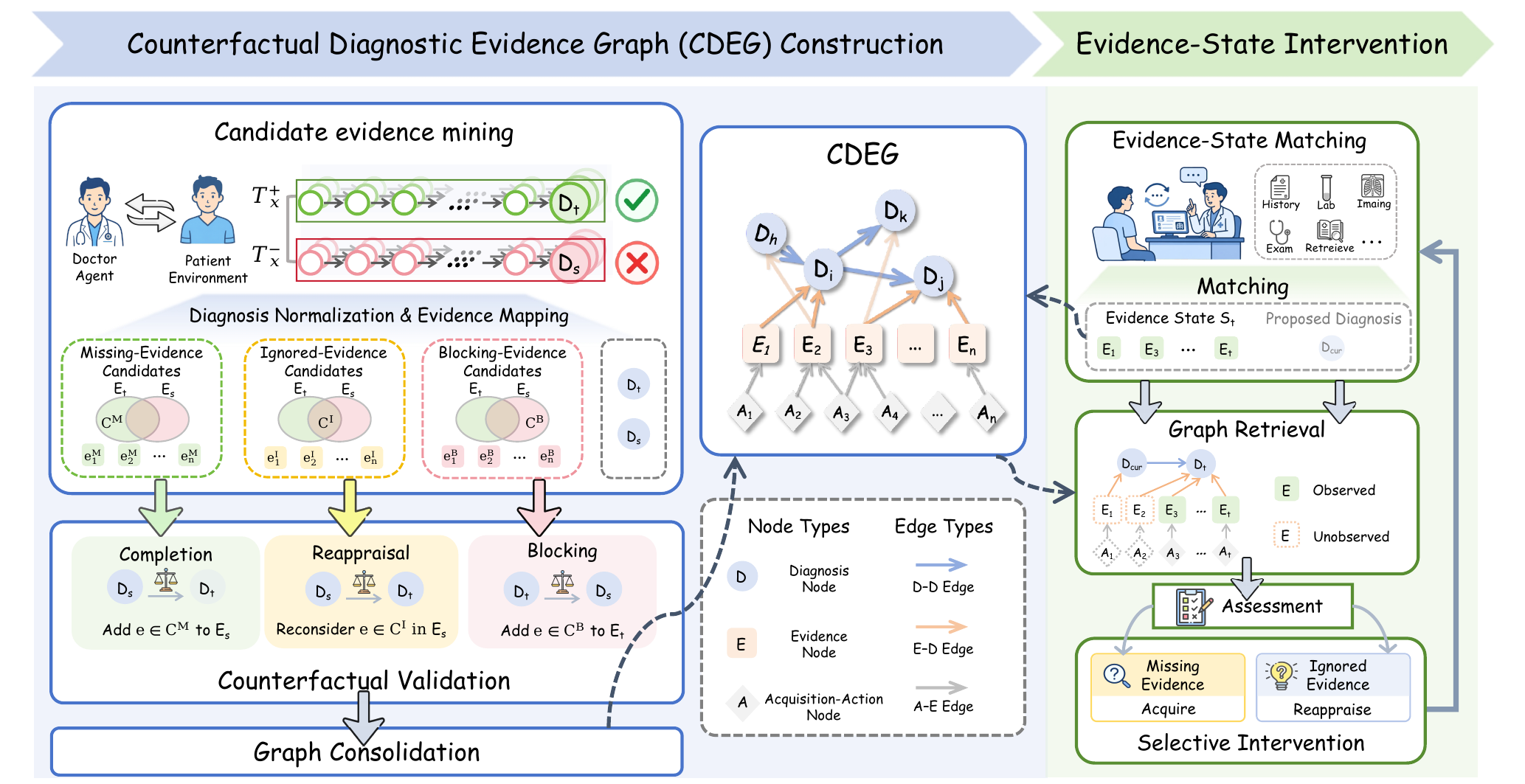}
    \caption{
    Overview of CDEG.
% CDEG compares successful and failed trajectories from the same case to identify evidence that may explain their different diagnoses. It tests whether adding missing evidence, reconsidering observed evidence, or introducing evidence against a revision changes diagnostic preference, and organizes validated evidence, diagnoses, and acquisition actions into a graph. 
CDEG first compares successful and failed trajectories from the same case to mine candidates that may explain their different diagnoses. It then tests whether adding missing evidence, reconsidering observed evidence, or introducing evidence against a revision changes diagnostic preference, and consolidates the validated evidence with diagnoses and acquisition actions into a graph. At inference time, CDEG retrieves relevant diagnostic edges based on the current evidence and guides the doctor agent to acquire missing evidence, reconsider ignored evidence, or avoid unnecessary intervention.
    }
    \label{fig:cdeg_method}
\end{figure*}

\section{Methods}

\subsection{Problem Setup}
\label{sec:problem_formulation}

Long-horizon interactive diagnosis models clinical diagnosis as a multi-turn sequential decision-making process under partial observability, in which a doctor agent progressively acquires and integrates clinical evidence before providing a final diagnosis~\cite{agentclinic,mediq}. Each clinical case is represented as $x=(\mathcal{E}_x,y_x^{*})$, where $\mathcal{E}_x$ denotes the patient environment and $y_x^{*}$ is the reference diagnosis. The environment contains the complete case information but initially reveals only $o_0$, with additional evidence becoming available through subsequent interaction.

At interaction turn $t$, the doctor agent conditions on the interaction history $h_t$ and selects its next action $a_t$:
\begin{equation}
h_t=\bigl(o_0,(a_i,o_i)_{i=1}^{t-1}\bigr),
\qquad
a_t\sim\pi_{\theta}(\cdot\mid h_t).
\end{equation}
where $\pi_{\theta}$ denotes the doctor agent policy. The agent may interact with the patient, use diagnostic tools to review clinical history, obtain laboratory or imaging evidence, perform examinations, retrieve relevant knowledge, or provide a final diagnosis. Each action other than the final diagnostic action yields an observation $o_t$, which is appended to the interaction history to form $h_{t+1}$ for the next turn.

The interaction terminates when the agent executes the final diagnostic action $a_T$ at turn $T$, yielding the trajectory:
\begin{equation}
\tau=\bigl(o_0,(a_t,o_t)_{t=1}^{T-1},a_T\bigr).
\end{equation}
Here, $a_T$ is the final diagnostic action, and $D(\tau)$ denotes the final diagnosis derived from the complete trajectory.

\subsection{Counterfactual Diagnostic Evidence Graph Construction}
\label{sec:cdeg_construction}
Figure~\ref{fig:cdeg_method} provides an overview of CDEG construction and evidence-state intervention.
CDEG construction proceeds in three stages. We first mine candidate evidence by comparing successful and failed trajectories from the same case, then validate each candidate by testing whether it changes diagnostic preference, and finally organize the validated evidence with its associated diagnoses and acquisition actions into the graph.

\paragraph{Candidate Evidence Mining.}

For each case $x$ in the graph-construction split, we sample $K$ independent rollouts of the doctor agent with the identical patient environment $\mathcal{E}_x$, yielding a set of trajectories $\mathcal{T}_x=\{\tau_x^{(k)}\}_{k=1}^{K}$. We map the final diagnosis from each rollout to the same ICD-aligned label space as the reference diagnosis and partition the trajectories by diagnostic correctness:
\begin{equation}
\begin{array}{l}
\mathcal{T}_x^{+}
= \{\tau \in \mathcal{T}_x \mid D(\tau)=y_x^{*}\}, \\[2pt]
\mathcal{T}_x^{-}
= \{\tau \in \mathcal{T}_x \mid D(\tau)\neq y_x^{*}\}.
\end{array}
\end{equation}
With the patient case fixed across rollouts, we compare successful and failed trajectories to uncover evidence acquisition and utilization patterns that distinguish correct from incorrect diagnoses.

However, trajectories from the same case may reach identical diagnoses through different interaction histories. We therefore pair each failed trajectory with the most similar successful trajectory based on the action--observation history $\tau_{\mathrm{pre}}$ before the final diagnostic action. For each failed trajectory $\tau^{-}$, we select
\begin{equation}
\tau^{+*}
=
\mathop{\mathrm{arg\,max}}_{\tau^{+}\in\mathcal{T}_x^{+}}
\cos\left(
\phi(\tau^{-}_{\mathrm{pre}}),
\phi(\tau^{+}_{\mathrm{pre}})
\right),
\end{equation}
where $\phi(\cdot)$ maps the action--observation history into a semantic embedding. The pair $(\tau^{-},\tau^{+*})$ is retained only when this maximum similarity exceeds a predefined threshold. For each retained pair, we denote their final diagnoses as $D_s$ and $D_t$, respectively.

We further transform heterogeneous clinical observations in each retained pair to
\emph{evidence atoms} grounded in UMLS concepts~\cite{bodenreider2004umls}. Each evidence atom represents a normalized clinical concept with its relevant attributes, including polarity, temporal status, numerical direction, and anatomical context. Observations mapped to the same evidence atom are merged while retaining their associated acquisition actions.

Let $E_s$ and $E_t$ denote the evidence atoms extracted from the failed and successful trajectories, respectively. Based on their relationships, We derive three candidate evidence sets:
\begin{equation}
\mathcal{C}^{M}=E_t\setminus E_s,
\qquad
\mathcal{C}^{I}=E_t\cap E_s,
\qquad
\mathcal{C}^{B}=E_s\setminus E_t.
\end{equation}
where $\mathcal{C}^{M}$, $\mathcal{C}^{I}$, and $\mathcal{C}^{B}$ denote missing-evidence, ignored-evidence, and blocking-evidence candidates, respectively. Evidence in $\mathcal{C}^{M}$ is acquired in the successful trajectory but absent from the failed trajectory, representing potential decision-critical evidence supporting the target diagnosis. Evidence in $\mathcal{C}^{I}$ appears in both trajectories, suggesting that the diagnostic discrepancy may stem from differences in evidence interpretation or utilization. Evidence in $\mathcal{C}^{B}$ appears only in the failed trajectory and may support the source diagnosis, indicating when revision toward the target diagnosis is unwarranted.

\paragraph{Counterfactual Validation.}

Building on these candidate sets, we next evaluate which one can alter diagnostic preference under a controlled counterfactual intervention. For each candidate, we apply only its corresponding intervention while keeping all other evidence unchanged.

Given an evidence set, the verifier compares $D_s$ and $D_t$ and selects the diagnosis better supported by the available evidence. We measure whether the verifier's preference changes after the intervention. To reduce order bias, all comparisons are repeated with the diagnosis order reversed, and a candidate is retained only if the same preference shift occurs in both orders. Accordingly, we perform different counterfactual interventions for the three candidate categories.

\textbf{(1) Completion test.}
For each $e\in\mathcal{C}^{M}$, we add $e$ to the failed-trajectory evidence $E_s$. If the verifier's preference changes from $D_s$ to $D_t$, $e$ is retained as validated missing evidence, indicating its potential role in supporting the target diagnosis.

\textbf{(2) Reappraisal test.}
For each $e\in\mathcal{C}^{I}$, we keep $E_s$ unchanged while explicitly exposing $e$ during diagnosis comparison. If the verifier's preference shifts from $D_s$ to $D_t$, $e$ is retained as validated ignored evidence, suggesting that the evidence was available but insufficiently utilized.

\textbf{(3) Blocking test.}
For each $e\in\mathcal{C}^{B}$, we add $e$ to the successful-trajectory evidence $E_t$. If the verifier's preference alters from $D_t$ to $D_s$, $e$ is retained as validated blocking evidence, indicating that revision toward the target diagnosis may be inappropriate when this evidence is present.

\paragraph{Graph Consolidation.}
We organize the counterfactually validated evidence, together with its associated diagnoses and acquisition actions, into the \emph{Counterfactual Diagnostic Evidence Graph}:
\begin{equation}
\mathcal{G}
=
\left(
\mathcal{V}_{D},
\mathcal{V}_{E},
\mathcal{V}_{A},
\mathcal{E}_{AE},
\mathcal{E}_{ED},
\mathcal{E}_{DD}
\right),
\end{equation}
where $\mathcal{V}_{D}$, $\mathcal{V}_{E}$, and $\mathcal{V}_{A}$ denote diagnosis, evidence, and acquisition-action nodes, respectively. $\mathcal{E}_{AE}$ maps acquisition actions to their resulting evidence, $\mathcal{E}_{ED}$ links evidence to the diagnoses it supports, and $\mathcal{E}_{DD}$ contains directed diagnostic-revision relations. Each $(D_s,D_t)\in\mathcal{E}_{DD}$ represents a revision from the diagnosis $D_s$ in a failed trajectory to the diagnosis $D_t$ in its paired successful trajectory for the same case. Such an edge is added in the graph only when counterfactual validation identifies evidence that shifts diagnostic preference from $D_s$ to $D_t$. Validated missing and ignored evidence are linked to $D_t$, while validated blocking evidence is connected to $D_s$. Therefore, each diagnostic revision relation retains both the evidence that promotes correction toward $D_t$ and the evidence that favors retaining $D_s$.

\subsection{Evidence-State Intervention}
\label{sec:online_intervention}

At inference time, CDEG maintains an evolving patient evidence state $\mathcal{S}_t$, which contains graph evidence nodes corresponding to the observations acquired up to interaction turn $t$. Newly obtained observations from the patient environment are mapped to graph evidence nodes through hybrid matching, which combines exact and semantic similarity matching. The resulting evidence nodes are added to $\mathcal{S}_t$ and used for future graph retrieval.

%这里retrieval和下面intervention看的有点迷惑。E_DD里，D_s对应的evidence不应该是容易带来误导的evidence嘛？有参考价值的往往是支持D_t的那些。所以能够支持一个正确诊断的E_DD，不是应该主要去参考D_t相关的evidence嘛？为什么要一起考虑?

\paragraph{Graph Retrieval.}
Given the current evidence state $\mathcal{S}_t$, CDEG retrieves the most related diagnosis-to-diagnosis edge. Specifically, for each $(D_s,D_t)\in\mathcal{E}_{DD}$, let $E_{s\rightarrow t}$ denote the evidence atoms associated with this edge. We measure the evidence coverage of the edge at turn $t$ as:
\begin{equation}
\mathrm{cov}_t(D_s,D_t)
=
\frac{
\left|\mathcal{S}_t\cap E_{s\rightarrow t}\right|
}{
\left|E_{s\rightarrow t}\right|
}.
\end{equation}
Before the doctor agent attempts to provide a final diagnosis, CDEG evaluates all edges in $\mathcal{E}_{DD}$ and retrieves the one with the highest coverage if its score exceeds a predefined threshold. Once the doctor agent attempts to provide a final diagnosis, the proposed diagnosis is matched to a graph diagnosis node $D_{\mathrm{cur}}$. CDEG then performs a localized retrieval over outgoing edges $(D_{\mathrm{cur}},D_t)\in\mathcal{E}_{DD}$ and selects the one with the highest coverage that exceeds the same threshold.

\paragraph{Selective Intervention.}

For the retrieved edge $(D_s,D_t)$, CDEG first determines whether the diagnostic revision is applicable to the current evidence state $\mathcal{S}_t$. If evidence supporting $D_s$ has already been observed, CDEG does not intervene along this edge.
Otherwise, CDEG checks whether any evidence associated with $D_t$ remains unobserved. If such evidence can be obtained through an available acquisition action, CDEG identifies a \emph{missing-evidence state} and prompts the doctor agent to perform the corresponding acquisition action to obtain the missing evidence.
If no further evidence can be acquired, CDEG considers diagnostic reappraisal when the doctor agent proposes $D_s$. It checks whether evidence associated with $D_t$ is already present in $\mathcal{S}_t$. CDEG identifies an \emph{ignored-evidence state} and prompts the doctor agent to reappraise the observed evidence before finalizing its diagnosis.

After each non-final action, newly acquired evidence is added to $\mathcal{S}_t$, and graph retrieval is repeated. When the doctor agent attempts to provide a final diagnosis, CDEG evaluates whether evidence acquisition or diagnostic reappraisal is required. If an intervention is triggered, the final diagnostic action is deferred and the interaction continues; otherwise, the proposed diagnosis is returned as the final output.

\section{Experiments}
\subsection{Experimental Setup}
\label{sec:experimental_setup}

\paragraph{Benchmarks.}
We evaluate CDEG on two long-horizon interactive diagnosis benchmarks. \emph{MIMIC-IDx} serves as the ID benchmark and is constructed by integrating MIMIC-IV~\cite{johnson2023mimic}, MIMIC-IV-Note~\cite{mimicivnote}, and matched MIMIC-CXR records~\cite{mimiccxr}. Each hospital admission is treated as a diagnostic case in which the patient environment initially exposes limited clinical information and additional evidence becomes available through subsequent interaction. Cases are partitioned at the patient level into disjoint graph-construction and evaluation sets. For OOD evaluation, we use the MedQA subset of \emph{AgentClinic}~\cite{agentclinic}, which converts MedQA cases~\cite{jin2021medqa} into text-based interactive diagnostic scenarios. The CDEG is constructed solely from the MIMIC-IDx graph-construction set, with the held-out MIMIC-IDx set and AgentClinic used exclusively for evaluation. Additional benchmark construction details are provided in the supplementary material, and benchmark statistics are summarized in Table~\ref{tab:dataset_statistics}.

\begin{table}[t]
\centering
\small
\setlength{\tabcolsep}{5.0pt}
\renewcommand{\arraystretch}{1.08}

\begin{tabular}{@{}lllr@{}}
\toprule
\textbf{Benchmark} &
\textbf{Source} &
\textbf{Setting} &
\textbf{\# Cases} \\
\midrule

\multirow{2}{*}{MIMIC-IDx}
& \multirow{2}{*}{MIMIC-IV}
& Graph Construction
& 573 \\

&
& ID Evaluation
& 200 \\

\midrule

AgentClinic
& MedQA
& OOD Evaluation
& 107 \\

\bottomrule
\end{tabular}

\caption{Benchmark statistics and experimental settings.}
\label{tab:dataset_statistics}
\end{table}

\paragraph{Baselines.}
We compare CDEG with the \emph{Vanilla Agent} and three baselines. \emph{Self-Reflection} prompts the doctor agent to reconsider its diagnostic reasoning from the current interaction history before providing a final diagnosis. \emph{Trajectory RAG} retrieves similar successful historical diagnostic trajectories as in-context examples. \emph{Flat Experience} retrieves diagnostic experience distilled from historical trajectories as unstructured text. For each backbone and benchmark, all methods use the same patient environment, available tools, and interaction budget. Historical information is restricted to the graph-construction split, and retrieval-based baselines use the same retrieval encoder and retrieval budget.

\begin{table*}[t]
\centering
\small
\setlength{\tabcolsep}{1.5mm}
\renewcommand{\arraystretch}{1.08}

\begin{tabular}{@{}ll*{3}{ccc}@{}}
\toprule
\multirow{2}{*}{\textbf{Backbone}} &
\multirow{2}{*}{\textbf{Method}} &
\multicolumn{3}{c}{\textbf{MIMIC-IDx (ID)}} &
\multicolumn{3}{c}{\textbf{AgentClinic (OOD)}} &
\multicolumn{3}{c}{\textbf{Avg.}} \\
\cmidrule(lr){3-5}
\cmidrule(lr){6-8}
\cmidrule(lr){9-11}

& &
\textbf{Acc. (\%)} & \textbf{Sim.} & \textbf{Corr. (\%)} &
\textbf{Acc. (\%)} & \textbf{Sim.} & \textbf{Corr. (\%)} &
\textbf{Acc. (\%)} & \textbf{Sim.} & \textbf{Corr. (\%)} \\
\midrule

% Gemini-3-Flash-Preview
\multirow{5}{*}{\shortstack[l]{Gemini-3-Flash-\\Preview}}
& Vanilla Agent
& 41.50 & 0.6584 & --
& 60.75 & 0.8442 & --
& 51.13 & 0.7513 & -- \\

& + Self-Reflection
& 47.00 & \underline{0.6828} & 23.08
& \underline{61.68} & \underline{0.8515} & 21.43
& 54.34 & \underline{0.7672} & 22.26 \\

& + Trajectory RAG
& 45.00 & 0.6726 & 22.22
& 60.75 & 0.7904 & 30.95
& 52.88 & 0.7315 & 26.59 \\

& + Flat Experience
& \underline{48.50} & 0.6688 & \underline{26.50}
& \underline{61.68} & 0.8161 & \underline{35.71}
& \underline{55.09} & 0.7425 & \underline{31.11} \\

& + \textbf{CDEG}
& \textbf{53.00} & \textbf{0.6908} & \textbf{33.33}
& \textbf{68.22} & \textbf{0.8576} & \textbf{45.24}
& \textbf{60.61} & \textbf{0.7742} & \textbf{39.29} \\
\midrule

% Gemini-3.1-Pro
\multirow{5}{*}{Gemini-3.1-Pro}
& Vanilla Agent
& 50.00 & 0.6733 & --
& 66.36 & 0.8691 & --
& 58.18 & 0.7712 & -- \\

& + Self-Reflection
& 47.00 & 0.6720 & 19.00
& 66.36 & 0.8674 & 25.00
& 56.68 & 0.7697 & 22.00 \\

& + Trajectory RAG
& \underline{52.50} & \underline{0.6846} & \underline{28.00}
& \underline{68.22} & 0.8690 & 30.56
& \underline{60.36} & \underline{0.7768} & \underline{29.28} \\

& + Flat Experience
& 50.00 & 0.6752 & 25.00
& 67.29 & \underline{0.8721} & \underline{33.33}
& 58.65 & 0.7737 & 29.17 \\

& + \textbf{CDEG}
& \textbf{59.00} & \textbf{0.7002} & \textbf{33.00}
& \textbf{71.96} & \textbf{0.8924} & \textbf{36.11}
& \textbf{65.48} & \textbf{0.7963} & \textbf{34.56} \\
\midrule

% GPT-5.6-Luna
\multirow{5}{*}{GPT-5.6-Luna}
& Vanilla Agent
& 50.00 & 0.6640 & --
& \underline{51.40} & \underline{0.7474} & --
& 50.70 & \underline{0.7057} & -- \\

& + Self-Reflection
& 43.00 & 0.6613 & 16.00
& 50.47 & 0.7397 & 17.31
& 46.74 & 0.7005 & 16.66 \\

& + Trajectory RAG
& 48.50 & 0.6718 & 22.00
& 47.66 & 0.7393 & 21.15
& 48.08 & 0.7056 & 21.58 \\

& + Flat Experience
& \underline{53.00} & \underline{0.6741} & \underline{24.00}
& 48.60 & 0.7340 & \underline{25.00}
& \underline{50.80} & 0.7041 & \underline{24.50} \\

& + \textbf{CDEG}
& \textbf{56.50} & \textbf{0.6942} & \textbf{33.00}
& \textbf{62.62} & \textbf{0.7952} & \textbf{36.54}
& \textbf{59.56} & \textbf{0.7447} & \textbf{34.77} \\
\midrule

% Claude-Sonnet-5
\multirow{5}{*}{Claude-Sonnet-5}
& Vanilla Agent
& 57.50 & 0.6666 & --
& 54.21 & \underline{0.7779} & --
& 55.86 & \underline{0.7223} & -- \\

& + Self-Reflection
& 58.00 & 0.6672 & 24.71
& 50.47 & 0.7531 & 24.49
& 54.24 & 0.7102 & 24.60 \\

& + Trajectory RAG
& \textbf{63.00} & \underline{0.6772} & \textbf{32.94}
& \underline{55.14} & 0.7663 & \underline{28.57}
& \underline{59.07} & 0.7218 & \underline{30.76} \\

& + Flat Experience
& 60.50 & 0.6669 & 29.41
& \underline{55.14} & 0.7629 & 22.45
& 57.82 & 0.7149 & 25.93 \\

& + \textbf{CDEG}
& \underline{61.00} & \textbf{0.7035} & \underline{30.59}
& \textbf{65.42} & \textbf{0.8331} & \textbf{46.94}
& \textbf{63.21} & \textbf{0.7683} & \textbf{38.77} \\
\bottomrule
\end{tabular}

\caption{
Diagnostic performance on MIMIC-IDx and AgentClinic.
Avg. denotes the macro-average over the two benchmarks.
The best and second-best results for each backbone are indicated by boldface and underlining.
}
\label{tab:main_results}
\end{table*}

\paragraph{Metrics.}
We primarily evaluate diagnostic quality using two complementary metrics, \emph{Accuracy} (Acc.) and \emph{Similarity} (Sim.). Accuracy measures the proportion of predicted diagnoses that are clinically consistent with the reference diagnoses, which is determined by GPT-5.6-Sol~\cite{openai2026gpt56} under a predefined clinical evaluation rubric~\cite{agentclinic}. Similarity captures the semantic proximity between predicted and reference diagnoses and is computed as cosine similarity in the embedding space of MedEmbed-large-v0.1~\cite{balachandran2024medembed}.

We further evaluate how each enhanced agent affects the decision behavior of the vanilla agent using \emph{Correction} (Corr.)~\cite{hassoon2026diagnostic} and \emph{Regression} (Reg.)~\cite{mintdiagnosis}. For each backbone, different methods are compared with the vanilla agent on the same cases. Let $N_{\mathrm{V}}^{-}$ and $N_{\mathrm{V}}^{+}$ denote the numbers of cases diagnosed incorrectly and correctly by the vanilla agent, respectively. Let $N_{\mathrm{fix}}$ denote the cases where the enhanced agent corrects errors made by the vanilla agent, and $N_{\mathrm{reg}}$ denote the cases where the enhanced agent changes correct predictions of the vanilla agent into incorrect ones. We define the two metrics as:
\begin{equation}
\mathrm{Correction}
=
\frac{N_{\mathrm{fix}}}{N_{\mathrm{V}}^{-}},
\qquad
\mathrm{Regression}
=
\frac{N_{\mathrm{reg}}}{N_{\mathrm{V}}^{+}}.
\end{equation}

\paragraph{Implementation Details.}

During graph construction, Gemini-3.1-Pro~\cite{google2026gemini31pro} and Gemini-3-Flash-Preview~\cite{google2025gemini3flash} were used to sample $K=4$ independent trajectories for each case in the MIMIC-IDx graph-construction split. Qwen3-Embedding-0.6B~\cite{zhang2025qwen3embedding} was used as the trajectory encoder and as the retrieval encoder for all retrieval-based baselines. Each failed trajectory was paired with its most similar successful trajectory, and pairs were retained only when their cosine similarity exceeded 0.7. GPT-5.5~\cite{openai2026gpt55} was used to map clinical observations to UMLS-grounded evidence atoms and to verify candidate evidence through counterfactual validation. The resulting CDEG was frozen and shared across all doctor-agent backbones. 

During inference, we compared Gemini-3-Flash-Preview~\cite{google2025gemini3flash}, Gemini-3.1-Pro~\cite{google2026gemini31pro}, GPT-5.6-Luna~\cite{openai2026gpt56}, and Claude-Sonnet-5~\cite{anthropic2026claudesonnet5} as doctor agent backbones. Newly acquired observations are mapped to graph evidence nodes through hybrid matching, which combines exact matching with SapBERT-based semantic matching using a similarity threshold of 0.72~\cite{liu2021sapbert}. Proposed diagnoses are mapped to graph diagnosis nodes using the same procedure with a threshold of 0.65. Diagnostic revision relations are retrieved when their evidence coverage exceeds 0.6. All methods use a maximum interaction budget of 30 turns. Additional prompts and implementation settings are provided in the supplementary material.

\subsection{Main Results}

As shown in Table~\ref{tab:main_results}, CDEG consistently improves diagnostic performance across all doctor agent backbones and both benchmarks. It increases the average accuracy across various backbones by 7.63\% on MIMIC-IDx and 8.88\% on AgentClinic. In contrast, Trajectory RAG and Flat Experience correct Vanilla Agent errors without producing consistent accuracy gains, suggesting that retrieving relevant historical experience alone is insufficient. The retained experience must also be reliable and applied under an appropriate evidence state. CDEG addresses both requirements by retaining evidence only when counterfactual intervention changes diagnostic preference and organizing the validated evidence into diagnostic revision relations for selective acquisition and reappraisal.
Additionally, although CDEG is constructed only from Gemini-3-Flash-Preview and Gemini-3.1-Pro-Preview trajectories, it still improves GPT-5.6-Luna and Claude-Sonnet-5 on both benchmarks. Its gains further transfer to the OOD evaluation set AgentClinic, suggesting that the validated evidence captures reusable diagnostic relations rather than reasoning patterns specific to a particular doctor-agent backbone or benchmark.

\subsection{Ablation Studies}

To further analyze the contributions of individual components, we conduct ablation studies on counterfactual validation for structured experience, different candidate evidence sets, evidence-state intervention, and diagnostic experience scaling. All studies use the MIMIC-IDx evaluation set with Gemini-3-Flash-Preview as the doctor-agent backbone.
\begin{table}[t]
\centering
\small
\setlength{\tabcolsep}{3.0pt}
\renewcommand{\arraystretch}{1.08}

\begin{tabular}{@{}lccc@{}}
\toprule
\textbf{Variant} &
\textbf{Acc. (\%)} &
\textbf{Sim.} &
\textbf{Corr. (\%)} \\
\midrule

\textbf{CDEG}
    & \textbf{53.00}
    & \textbf{0.6908}
    & \textbf{33.33} \\

\midrule
\multicolumn{4}{@{}l}{\textit{Counterfactual Validation for Structured Experience}} \\
Flat Experience
    & \underline{48.50}
    & \underline{0.6688}
    & \underline{26.50} \\
w/o Counterfactual Validation
    & 43.50
    & 0.6481
    & 24.79 \\

\midrule
\multicolumn{4}{@{}l}{\textit{Different Candidate Evidence Sets}} \\
w/o Missing Evidence
    & 51.00
    & 0.6865
    & 25.64 \\
w/o Ignored Evidence
    & 49.00
    & 0.6875
    & 25.64 \\
w/o Blocking Evidence
    & \underline{51.50}
    & \underline{0.6904}
    & \underline{30.77} \\

\midrule
\multicolumn{4}{@{}l}{\textit{Evidence Selective Intervention}} \\
Acquisition Only
    & 43.50
    & 0.6702
    & 19.66 \\
Reappraisal Only
    & \underline{47.50}
    & \underline{0.6769}
    & \underline{29.91} \\

\bottomrule
\end{tabular}

\caption{
Ablation studies of CDEG on MIMIC-IDx using Gemini-3-Flash-Preview.
CDEG is the full-model reference, and the best variant within each ablation group is underlined.
}
\label{tab:cdeg_ablation}
\end{table}

\subsubsection{Counterfactual Validation for Structured Experience}

As shown in Table~\ref{tab:cdeg_ablation}, removing counterfactual validation reduces accuracy from 53.00\% to 43.50\% and even performs worse than Flat Experience at 48.50\%. This result shows that CDEG's gains do not arise from graph structuring alone. The graph organizes trajectory-derived experience into structured diagnostic relations for efficient storage and retrieval, while the key advantage of CDEG lies in using counterfactual validation to determine which candidate evidence should be consolidated into the graph. By retaining only evidence that changes diagnostic preference, CDEG excludes noisy or incidental trajectory differences and concentrates diagnostically useful information in the graph.

\subsubsection{Different Candidate Evidence Sets}

As shown in Table~\ref{tab:cdeg_ablation}, removing either missing or ignored evidence reduces Correction from 33.33\% to 25.64\%, indicating that both are important for broadening the coverage of decision-critical evidence represented in the graph. Without either, less validated evidence is available across different evidence states, weakening CDEG's guidance for diagnostic correction. Removing blocking evidence causes a smaller decline to 30.77\%, suggesting that it helps CDEG recognize the boundary conditions of a diagnostic revision and reduce inappropriate interventions under incompatible evidence states.

\subsubsection{Evidence Selective Intervention}
As shown in Table~\ref{tab:cdeg_ablation}, Reappraisal Only improves Correction by 10.25\% over Acquisition Only, yet both underperform the complete CDEG. This pattern indicates that long-horizon diagnostic errors cannot be fully addressed by a single intervention. By dynamically selecting between acquisition and reappraisal according to the current evidence state, CDEG can capture correction opportunities missed by either intervention alone.

\subsubsection{Diagnostic Experience Scaling}

\begin{figure}[t]
    \centering
    \includegraphics[width=\columnwidth]{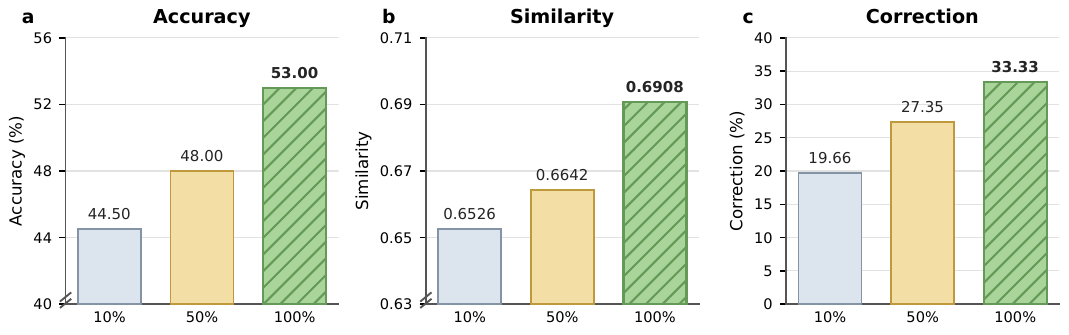}
    \caption{Effect of diagnostic experience scaling.}
    \label{fig:experience_scaling}
\end{figure}

As shown in Figure~\ref{fig:experience_scaling}, increasing graph-construction data from 10\% to 100\% raises Correction from 19.66\% to 33.33\%, with Accuracy and Similarity improving in parallel. This trend suggests that additional diagnostic trajectories broaden the coverage of counterfactually validated diagnostic revision relations, allowing CDEG to provide relevant evidence guidance for a wider range of subsequent cases.

\subsection{Analysis}
\subsubsection{Evidence States at Diagnostic Failure}

We audit 117 incorrect trajectories produced by the vanilla agent on MIMIC-IDx, and categorize each trajectory by its evidence sufficiency before final diagnostic commitment. The audit protocol is provided in the supplementary material. As shown in Table~\ref{tab:evidence_sufficiency}, only 11.97\% of errors occur under conflicting evidence. In contrast, 50.43\% arise from incomplete decision-critical evidence acquisition, while another 37.60\% occur despite such evidence already being available but insufficiently incorporated into the diagnosis. Together, these two failure modes account for 88.03\% of all errors, showing that reliable long-horizon diagnosis hinges on effective management of decision-critical evidence across acquisition and use. CDEG explicitly addresses both failures by using the current evidence state to dynamically guide the doctor agent to acquire missing evidence or reappraise overlooked evidence.

\begin{table}[t]
\centering
\small
\setlength{\tabcolsep}{6.0pt}
\renewcommand{\arraystretch}{1.08}

\begin{tabular}{@{}lrr@{}}
\toprule
\textbf{Evidence Sufficiency} &
\textbf{\# Cases} &
\textbf{Errors (\%)} \\
\midrule
Incomplete  & 59 & 50.43 \\
Sufficient  & 44 & 37.60 \\
Conflicting & 14 & 11.97 \\
\midrule
Total       & 117 & 100.00 \\
\bottomrule
\end{tabular}

\caption{
Evidence states among incorrect trajectories produced by vanilla agent on MIMIC-IDx.
}
\label{tab:evidence_sufficiency}
\end{table}

\subsubsection{Evidence-State Intervention and Diagnostic Stability}

\begin{figure}[t]
    \centering
    \includegraphics[width=\columnwidth]{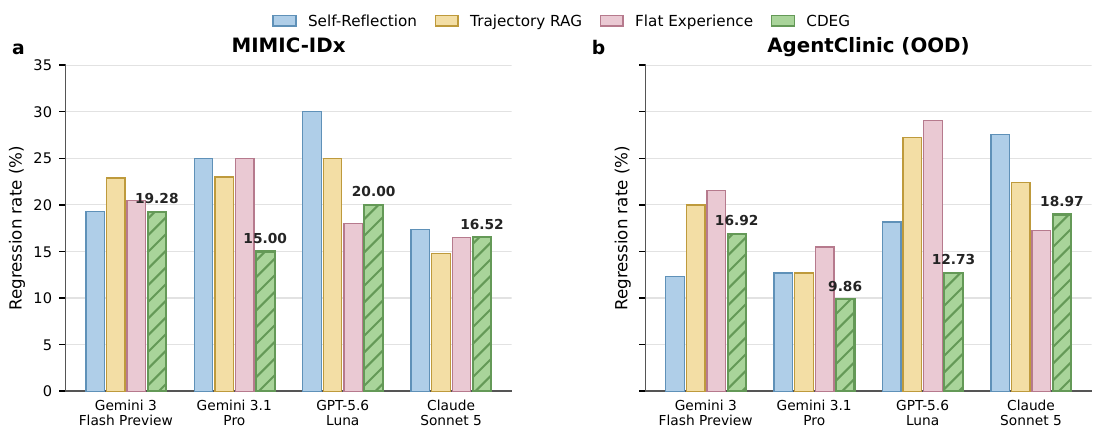}
    \caption{
Regression rates across doctor-agent backbones on MIMIC-IDx and AgentClinic. 
}

    \label{fig:cdeg_regression}
\end{figure}

As shown in Figure~\ref{fig:cdeg_regression}, CDEG achieves the lowest average Regression among the evaluated methods, with 17.70\% on MIMIC-IDx and 14.62\% on AgentClinic, while maintaining higher Correction exhibited in Table~\ref{tab:main_results}. This combination indicates that CDEG achieves a better balance between correcting diagnostic errors and preserving already correct diagnoses, rather than simply increasing the frequency of revision. This pattern is consistent with evidence-state intervention, which determines whether a retrieved diagnostic revision is applicable to the current evidence state before triggering acquisition or reappraisal.

\subsubsection{Diagnostic Gains Beyond Longer Interaction}

\begin{figure}[t]
    \centering
    \includegraphics[width=\columnwidth]{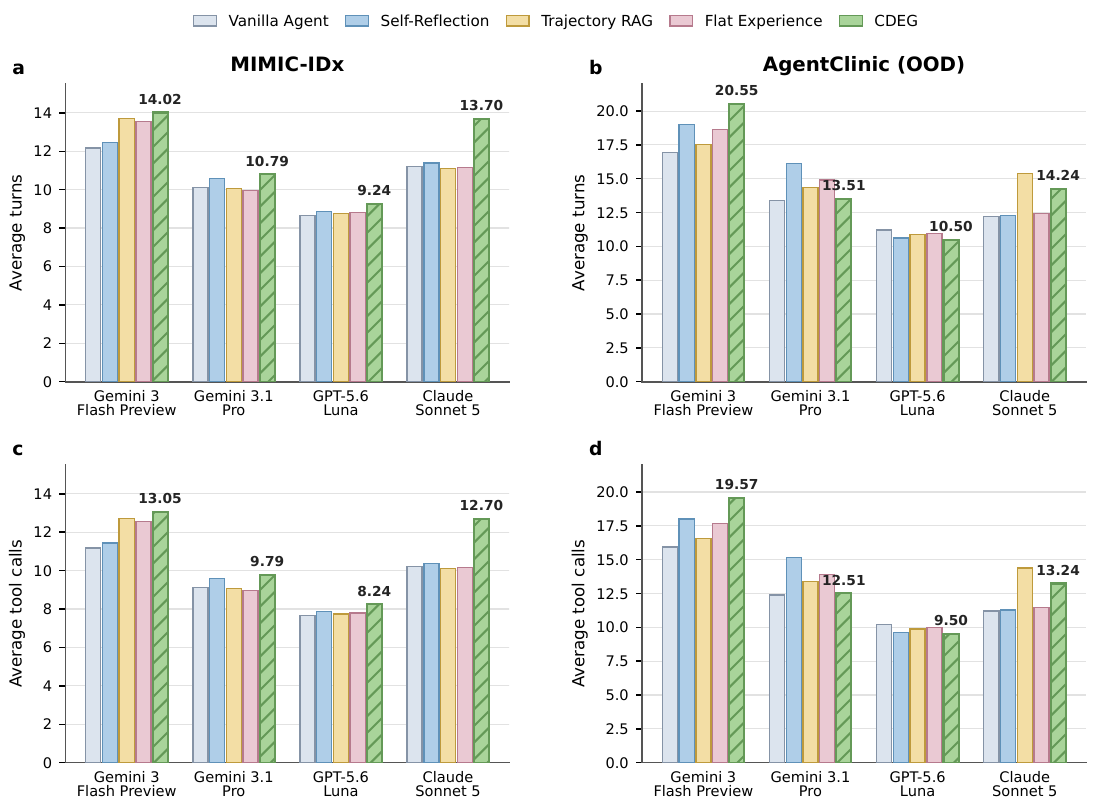}
    \caption{Interaction turns and tool-call counts across doctor-agent backbones on MIMIC-IDx and AgentClinic.}
    \label{fig:cdeg_turn_tool}
\end{figure}

As shown in Figure~\ref{fig:cdeg_turn_tool}, CDEG improves diagnostic performance with only 1.41 and 1.27 additional turns on MIMIC-IDx and AgentClinic, indicating more efficient use of the interaction budget. Rather than indiscriminately extending the diagnostic process, CDEG uses the evidence state to selectively guide acquisition of missing evidence or reappraisal of observed evidence. This effect is particularly apparent for the GPT-5.6-Luna backbone on AgentClinic, where CDEG improves Accuracy by 11.22\% while reducing the average interaction length by 0.71 turns, suggesting that it reduces ineffective exploration and reaches an evidence state sufficient for diagnosis more quickly. Tool-call counts exhibit the same pattern, further demonstrating that CDEG improves interaction efficiency through targeted evidence acquisition or reappraisal.

\section{Conclusion}

We presented CDEG, a graph-based framework that learns reusable decision-critical evidence by contrasting successful and failed diagnostic trajectories, validating candidate evidence through controlled counterfactual interventions, and organizing validated evidence, diagnoses, and actions into a structured graph that guides acquisition or reappraisal according to the evolving patient evidence state. Across ID and OOD benchmarks with four doctor-agent backbones, CDEG consistently improves vanilla agent diagnostic performance, and achieves the best performance. In the future, we will explore online graph evolution for continual experience validation and extend CDEG to broader clinical scenarios.

\bibliography{aaai2027}

\appendix

% =========================================================
% A. Experimental Details
% =========================================================
\section{Experimental Details}
\label{app:experimental_details}

\subsection{Benchmark Construction and Data Splits}
\label{app:benchmark_construction}

\paragraph{MIMIC-IDx.}
MIMIC-IDx is constructed by integrating MIMIC-IV, MIMIC-IV-Note, and matched MIMIC-CXR records. Each hospital admission is treated as one long-horizon interactive diagnostic case. To approximate the information available during diagnosis before final diagnostic commitment, we retain records from the pre-treatment diagnostic phase and remove content that directly reveals the reference diagnosis. Each case is instantiated as a patient environment that initially exposes limited clinical information through an initial observation, while additional patient-specific clinical evidence can be progressively acquired through interaction. The available evidence includes demographics, clinical history, physical examination findings, laboratory and microbiology results, diagnostic procedures and their findings, and radiology reports.

We partition the cases at the patient level into two disjoint splits, using 573 cases to generate trajectories for CDEG construction and reserving 200 cases exclusively for held-out in-domain evaluation. Reference diagnoses are not exposed to the doctor agent during interaction. They are used only to determine trajectory correctness during graph construction and for held-out diagnostic evaluation and post hoc evidence-sufficiency analysis.

\paragraph{AgentClinic.}
For out-of-distribution evaluation, we use the MedQA subset of AgentClinic, which converts 107 MedQA cases into text-based interactive diagnostic scenarios. Compared with MIMIC-IDx, AgentClinic differs in its underlying data source and diagnostic distribution. It is used exclusively as a held-out out-of-distribution evaluation benchmark and is never accessed during CDEG construction.

\subsection{Diagnostic Environment and Tool Interface}
\label{app:diagnostic_environment}

Across both benchmarks, the doctor agent interacts with the patient environment through a common action interface. At each turn, the agent selects one action. Each non-final action returns an observation determined by the information available in the underlying case. When the requested patient-specific information is unavailable, the environment returns a standardized unavailable-information response. The available actions are defined as follows.

\paragraph{\texttt{ask\_patient(question)}.}
Queries one patient-specific item of clinical history, such as a current symptom, its onset or time course, medication use, a relevant exposure, or past medical history.

\paragraph{\texttt{request\_exam(exam\_type)}.}
Requests one patient-specific clinical observation from physical examination or structured diagnostic records, including vital signs, laboratory or microbiology results, and findings from completed diagnostic procedures.

\paragraph{\texttt{cxr()}.}
Requests chest-imaging evidence available in the current case. For MIMIC-IDx, this evidence is obtained from the matched MIMIC-CXR record and its associated radiology report when available.

\paragraph{\texttt{retrieve(query)}.}
Retrieves general medical knowledge relevant to the current diagnostic problem, such as diagnostic criteria, differential diagnoses, or potentially informative examinations.

\paragraph{\texttt{diagnosis(result)}.}
Issues the final diagnostic action with the diagnosis and terminates the interaction.

\subsection{Generation and Inference Settings}
\label{app:generation_settings}

Doctor-agent actions are generated with temperature $0.4$ and top-$p=0.95$, subject to a maximum budget of 30 interaction turns. Within each backbone--benchmark setting, all methods use the same doctor-agent decoding configuration, patient environment, available tools, and interaction budget.

Auxiliary LLM calls for evidence normalization, counterfactual validation, self-reflection, and diagnostic evaluation use temperature $0$. Trajectory RAG and Flat Experience each retrieve one item from the graph-construction split using the shared retrieval encoder. Any acquisition action triggered by CDEG remains subject to the shared 30-turn interaction budget.

\begin{table*}[t]
\centering
\small
\setlength{\tabcolsep}{5.0pt}
\renewcommand{\arraystretch}{1.08}
\begin{tabular}{@{}
    >{\raggedright\arraybackslash}p{0.25\textwidth}
    >{\centering\arraybackslash}p{0.10\textwidth}
    >{\raggedright\arraybackslash}p{0.57\textwidth}
    @{}}
\toprule
\textbf{Hyperparameter} & \textbf{Value} & \textbf{Description} \\
\midrule

\multicolumn{3}{@{}l}{\textit{Trajectory Generation and Candidate Evidence Mining}} \\

Trajectories per case ($K$)
& 4
& Number of independent diagnostic trajectories generated for each
MIMIC-IDx graph-construction case using the two construction backbones. \\

Trajectory-pair similarity threshold
& 0.70
& Minimum cosine similarity for retaining a failed trajectory and its
most similar successful trajectory from the same case. \\

Evidence-normalization temperature
& 0
& Temperature used for schema-constrained evidence-atom normalization
and UMLS alignment. \\

\midrule
\multicolumn{3}{@{}l}{\textit{Counterfactual Validation}} \\

Counterfactual-verifier temperature
& 0
& Temperature used for the completion, reappraisal, and blocking tests. \\

\midrule
\multicolumn{3}{@{}l}{\textit{Evidence-State Intervention}} \\

Evidence-matching threshold
& 0.72
& Minimum SapBERT similarity for semantically matching an observation
to a graph evidence node when no exact match is found. \\

Diagnosis-matching threshold
& 0.65
& Minimum SapBERT similarity for matching a proposed diagnosis to a
graph diagnosis node. \\

Diagnostic-relation coverage threshold
& 0.60
& Minimum evidence coverage required for retrieving a diagnostic
revision relation. \\

\midrule
\multicolumn{3}{@{}l}{\textit{Baseline Implementation}} \\

Retrieval budget (top-$k$)
& 1
& Number of trajectories or flat experience entries retrieved by
Trajectory RAG and Flat Experience. \\

\midrule
\multicolumn{3}{@{}l}{\textit{Doctor-Agent Generation}} \\

Doctor-agent temperature
& 0.4
& Temperature used for doctor-agent action generation during trajectory
generation and evaluation. \\

Doctor-agent top-$p$
& 0.95
& Nucleus-sampling parameter used for doctor-agent action generation. \\

Maximum output tokens
& 8,192
& Maximum output-token budget for each doctor-agent action-generation
call. \\

Maximum interaction turns
& 30
& Maximum doctor--environment interaction budget for each case. \\

\midrule
\multicolumn{3}{@{}l}{\textit{Diagnostic Evaluation and Evidence-Sufficiency Audit}} \\

Diagnostic-evaluator temperature
& 0
& Temperature used by the fixed evaluator under the shared diagnostic evaluation rubric. \\

Evidence-audit temperature
& 0
& Temperature used in both passes of the evidence-sufficiency audit. \\

\bottomrule
\end{tabular}
\caption{Principal hyperparameters used for CDEG construction,
evidence-state intervention, baseline implementation, doctor-agent
generation, and evaluation.}
\label{tab:app_hyperparameters}
\end{table*}

\subsection{Baseline Implementations}
\label{app:baseline_details}

\paragraph{Self-Reflection.}
Before providing a final diagnosis, Self-Reflection introduces a single reflection step using the same doctor-agent backbone. Given only the current interaction history and the proposed diagnosis, the agent reassesses whether the diagnosis is adequately supported and whether any observed evidence has been insufficiently incorporated into its diagnostic reasoning.

\paragraph{Trajectory RAG.}
Trajectory RAG constructs a retrieval pool from successful historical diagnostic trajectories in the graph-construction split. Using the interaction context available to the doctor agent at retrieval time as the query, Qwen3-Embedding-0.6B retrieves the top-1 most similar trajectory. The retrieved action--observation history and final diagnosis are provided to the doctor agent as an in-context example.

\paragraph{Flat Experience.}
Flat Experience distills historical diagnostic trajectories into unstructured natural-language experience entries. The retrieval pool contains entries distilled either from individual successful trajectories or from paired successful and failed trajectories from the same case. Each entry summarizes the transferable clinical context, relevant evidence, diagnostic lesson, resulting diagnosis, and conditions under which the experience may apply. At inference time, Qwen3-Embedding-0.6B uses the interaction context available to the doctor agent to retrieve the top-1 most similar entry, which is provided as additional context. 

\subsection{Evidence-Sufficiency Audit Procedure}
\label{app:evidence_audit}

We audit the 117 incorrect trajectories produced by the Gemini-3-Flash-Preview vanilla agent on the MIMIC-IDx evaluation split. For each trajectory, the audit examines all patient-specific clinical evidence observed before the final diagnostic
action and characterizes the evidence state at diagnostic failure. The audit distinguishes errors arising from incomplete decision-critical evidence acquisition, insufficient incorporation of evidence already observed, and unresolved competition between diagnostic interpretations.

A fixed GPT-5.5 adjudicator performs the audit in two passes. In the first pass, a blind evidence parser receives the action--observation history preceding the final diagnostic action and records the patient-specific evidence observed by the doctor agent. It does not receive either the reference diagnosis or the vanilla agent's
final diagnosis, preventing diagnostic labels from influencing the evidence parse. In the second pass, an evidence-sufficiency adjudicator receives the blind evidence parse, the original pre-final action--observation history, the vanilla agent diagnosis, and the reference diagnosis. It identifies the observed evidence supporting the reference diagnosis, key discriminative evidence that remains unobserved, and observed evidence supporting the vanilla agent diagnosis or other competing diagnoses. Based on these
findings, it assigns each trajectory to exactly one of the following evidence states:

\begin{itemize}
\item \textbf{Incomplete}: The evidence observed before the final diagnostic action is insufficient to distinguish the reference diagnosis from competing diagnoses, and key discriminative evidence remains unobserved. The error therefore primarily reflects incomplete decision-critical evidence acquisition.

\item \textbf{Sufficient}: The observed evidence is sufficient to favor the reference diagnosis over the vanilla agent diagnosis without requiring additional patient-specific evidence. The error therefore primarily reflects insufficient incorporation of evidence already available before the final diagnostic action.

\item \textbf{Conflicting}: The observed evidence materially supports competing diagnostic interpretations and does not reliably resolve the diagnostic boundary between them.
\end{itemize}

\subsection{Hyperparameter Details}
\label{app:hyperparameter_details}

Table~\ref{tab:app_hyperparameters} summarizes the principal hyperparameters used for CDEG construction, evidence-state intervention, baseline implementation, doctor-agent generation, diagnostic evaluation, and evidence-sufficiency auditing. These settings are fixed across cases and evaluation runs. Within each backbone--benchmark setting, all methods use the same doctor-agent generation settings and interaction budget.

% =========================================================
% B. Method Details
% =========================================================
\section{Method Details}
\label{app:cdeg_implementation}

\subsection{Trajectory Pairing and Evidence Normalization}
\label{app:candidate_mining_details}

For each case in the graph-construction split, we generate $K=4$ independent diagnostic trajectories using Gemini-3-Flash-Preview and Gemini-3.1-Pro as the construction backbones, with two rollouts from each backbone.

For each failed trajectory $\tau^{-}$, Qwen3-Embedding-0.6B encodes its action--observation history before the final diagnostic action and those of all successful trajectories from the same case. We select the most similar successful trajectory $\tau^{+*}$ by cosine similarity and retain the pair $(\tau^{-},\tau^{+*})$ when its similarity meets the threshold reported in Table~\ref{tab:app_hyperparameters}.

After trajectory pairing, clinical evidence mentions in each retained pair are normalized into UMLS-grounded evidence atoms using GPT-5.5 with a fixed schema-constrained prompt. The normalization step maps each supplied evidence mention to a canonical clinical concept and preserves its assertion polarity, temporal status, numerical direction when applicable, and anatomical context. Mentions mapped to the same evidence atom are merged while retaining their associated acquisition actions and trajectory provenance. Final diagnoses are separately normalized to the ICD-aligned diagnosis space.

The normalized evidence sets of each retained pair are then contrasted to construct the missing-evidence, ignored-evidence, and blocking-evidence candidate sets.

\subsection{CDEG Construction Statistics}
\label{app:cdeg_statistics}

\paragraph{Construction and graph statistics.}
As shown in Table~\ref{tab:app_graph_statistics}, CDEG progressively transforms paired successful and failed trajectories into candidate evidence and then consolidates the validated evidence into structured diagnosis--evidence--action relations. These statistics characterize how trajectory-level experience is converted into reusable evidence-level guidance.

\begin{table}[t]
\centering
\small
\setlength{\tabcolsep}{4.2pt}
\renewcommand{\arraystretch}{1.06}
\begin{tabular}{@{}lr@{}}
\toprule
\textbf{Statistic} & \textbf{Count} \\
\midrule

\multicolumn{2}{@{}l}{\textit{Trajectory generation and pairing}} \\
Graph-construction cases & 573 \\
Diagnostic trajectories & 2,292 \\
Cases with $\mathcal{T}_x^{+}$ and $\mathcal{T}_x^{-}$ & 461 \\

\midrule
\multicolumn{2}{@{}l}{\textit{Candidate evidence before validation}} \\
Missing-evidence candidate instances & 971 \\
Ignored-evidence candidate instances & 843 \\
Blocking-evidence candidate instances & 457 \\
Total candidate evidence instances & 2,271 \\

\midrule
\multicolumn{2}{@{}l}{\textit{Frozen CDEG after graph consolidation}} \\
Diagnosis nodes $\lvert\mathcal{V}_{D}\rvert$ & 437 \\
Evidence nodes $\lvert\mathcal{V}_{E}\rvert$ & 516 \\
Acquisition-action nodes $\lvert\mathcal{V}_{A}\rvert$ & 203 \\
Total graph nodes & 1,156 \\
Diagnosis--diagnosis relations $\lvert\mathcal{E}_{DD}\rvert$ & 270 \\
Evidence--diagnosis relations $\lvert\mathcal{E}_{ED}\rvert$ & 692 \\
Action--evidence relations $\lvert\mathcal{E}_{AE}\rvert$ & 464 \\
Total graph relations & 1,426 \\

\bottomrule
\end{tabular}
\caption{CDEG construction statistics. Candidate evidence instances are
counted before counterfactual validation at the trajectory-pair level,
whereas graph nodes and relations are counted after counterfactual
validation and graph consolidation.}
\label{tab:app_graph_statistics}
\end{table}

\paragraph{Counterfactual validation outcomes.}
As shown in Table~\ref{tab:app_validation_statistics}, counterfactual validation retains 30.47\% of the mined candidate evidence, with different retention patterns across missing, ignored, and blocking candidates. This selectivity indicates that trajectory contrasts provide broad but noisy candidates across distinct evidence
roles, further supporting the need for counterfactual validation before graph consolidation.
\begin{table}[t]
\centering
\small
\setlength{\tabcolsep}{4.2pt}
\renewcommand{\arraystretch}{1.06}
\begin{tabular}{@{}lrrr@{}}
\toprule
\textbf{Candidate Type} &
\textbf{Mined} &
\textbf{Validated} &
\textbf{Retention (\%)} \\
\midrule

Missing-evidence candidates
& 971
& 342
& 35.22 \\

Ignored-evidence candidates
& 843
& 269
& 31.91 \\

Blocking-evidence candidates
& 457
& 81
& 17.72 \\

\midrule
Total
& 2,271
& 692
& 30.47 \\

\bottomrule
\end{tabular}
\caption{Counterfactual validation statistics before graph
consolidation. Retention denotes the proportion of candidate evidence
instances satisfying the required diagnostic preference shift under both
diagnosis orders.}
\label{tab:app_validation_statistics}
\end{table}

% =========================================================
% C. Prompt Templates
% =========================================================
\section{Prompt Templates}
\label{app:prompts}

\subsection{Agent and Baseline Prompts}
\label{app:doctor_prompt}

\prompttitle{Doctor Agent}
\begin{lstlisting}[style=prompt]
You are a doctor conducting a multi-turn clinical interaction to determine the patient's primary diagnosis.

At each turn, select exactly one action from the available tool schema. Do not return plain text outside the selected tool call.

Available actions:
- ask_patient(question): ask one focused question about symptoms, history, disease course, medications, or exposures.
- request_exam(exam_type): request a patient-specific physical examination, vital sign, laboratory result, or other available clinical finding.
- cxr(): request chest-imaging report available in the current case.
- retrieve(query): retrieve general medical knowledge relevant to diagnostic reasoning.
- diagnosis(result): submit one proposed primary diagnosis.

Use the complete interaction history to avoid redundant actions and select the action expected to provide the most diagnostically informative patient-specific evidence. Only patient or tool observations returned by the diagnostic environment may be treated as evidence about the current patient. Retrieved knowledge is background information and must not be treated as a patient-specific observation. Do not infer a positive, negative, or normal finding from a diagnostic criterion, recommended examination, pending test, or unavailable result.

When sufficient patient-specific evidence has been collected, call diagnosis and format its result argument as:
The final diagnosis is: \boxed{[DIAGNOSIS]}.
\end{lstlisting}

\prompttitle{Self-Reflection}
\label{app:self_reflection_prompt}

\begin{lstlisting}[style=prompt]
You are a doctor performing one final self-reflection before submitting the primary diagnosis. Reassess the current draft using the evidence acquired during the current interaction. 

Decision procedure:
1. Reconstruct a compact evidence summary covering the chief presentation, decisive positive findings, decisive negative findings, and time course.
2. Identify which decisive findings support the draft and which remain unexplained.
3. Identify the strongest observed evidence that contradicts the draft or represents a major presenting finding that the draft cannot explain. An unrelated finding is not automatically contradictory, and missing information is not negative evidence.
4. Compare the draft with the single strongest competing primary diagnosis.
5. Check whether each diagnosis explains the chief presentation, time course, and decisive findings at the specificity supported by the evidence.
6. Revise only when the alternative explains more decisive observed evidence with fewer major contradictions and is therefore clearly better supported as the primary diagnosis.
7. A comorbidity, incidental finding, or merely possible alternative is insufficient to replace the primary diagnosis.

Constraints:
- Do not invent evidence or request additional patient-specific information.
- Cite only findings observed in the current interaction.
- Perform this reflection once and return one final diagnosis label.

[CURRENT INTERACTION HISTORY]
[CURRENT DRAFT DIAGNOSIS]

Return JSON only:
{
  "decision": "KEEP_DRAFT|REVISE",
  "final_diagnosis": "..."
}
\end{lstlisting}

\prompttitle{Trajectory RAG}
\label{app:trajectory_rag_prompt}

\begin{lstlisting}[style=prompt]
You are a doctor conducting a multi-turn clinical interaction. One successful diagnostic trajectory from a different patient is provided as an in-context example. Follow the shared doctor-agent instructions and select exactly one action from the available tool schema at each turn.

Rules:
1. The retrieved trajectory is from a different patient. Its observations are not evidence about the current patient.
2. Transfer only clinically applicable guidance, including which evidence was acquired, which findings distinguished competing diagnoses, and how the final diagnosis was supported.
3. Base every diagnostic decision on the current patient's interaction history and patient-specific observations. Retrieved general knowledge may support reasoning but must not be treated as evidence about the current patient.
4. Ignore the retrieved trajectory when it is not applicable. Do not invent positive, negative, normal, or unavailable findings for the current patient.
5. When sufficient patient-specific evidence has been acquired, submit one primary diagnosis at the specificity supported by the available evidence.

[CURRENT INTERACTION HISTORY]
[RETRIEVED TRAJECTORY]

Return exactly one tool call and no additional text.
\end{lstlisting}

\prompttitle{Flat Experience Distillation}
\label{app:flat_extraction_prompt}

\begin{lstlisting}[style=prompt]
You are a clinical experience curator for a diagnostic agent. Derive one transferable diagnostic experience from the supplied historical trajectory source.

The source contains either:
- one successful diagnostic trajectory; or
- one successful trajectory paired with one failed trajectory from the same case.

Use only information explicitly supported by the supplied trajectory source. When a paired failed trajectory is provided, you may use grounded differences between the successful and failed trajectories to identify transferable evidence-acquisition or diagnostic-reasoning lessons. Do not infer that a finding is absent merely because it is not mentioned in one trajectory.

Treat retrieved general knowledge as background context rather than patient-specific evidence. Represent unavailable or unobserved information as unknown rather than positive or negative. Ensure that every referenced finding, test, action, diagnosis, and turn number is grounded in the supplied trajectory source.

Create at most one transferable FLAT EXPERIENCE entry as a self-contained paragraph of unstructured natural-language guidance. Organize the paragraph according to the following reasoning flow:
relevant clinical context -> clinically relevant evidence -> diagnostic reasoning -> supported diagnosis and applicability conditions.

Rules:
1. The experience should be applicable to a future diagnostic interaction while remaining independent of its source case. When further evidence would be necessary, identify the category of patient-specific evidence that should be verified.

2. Generalize incidental details such as exact age, wording, chronology, and numerical values unless they materially affect the diagnostic distinction. Present the lesson as general clinical guidance rather than as a description of a trajectory, agent, memory, source case, or sequence of turns.

3. Diagnoses may be named when necessary to express a clinically meaningful distinction, but the entry should explain the supporting evidence and reasoning rather than merely restating the successful final diagnosis.

4. When a failed trajectory is supplied, use it only to identify grounded contrasts in acquired evidence, evidence interpretation, or diagnostic reasoning. Do not treat the failed trajectory as evidence that its final diagnosis is clinically valid.

5. State a concrete applicability boundary by identifying evidence that would make the lesson inapplicable or require a different diagnostic interpretation.

6. Do not include treatment recommendations. Keep the experience between 50 and 100 English words.

[HISTORICAL SUCCESSFUL DIAGNOSTIC TRAJECTORY]

[OPTIONAL PAIRED FAILED DIAGNOSTIC TRAJECTORY;
OMITTED WHEN NO PAIRED FAILED TRAJECTORY IS USED]

Return JSON only:
{
  "experiences": [
    {
      "text": "one self-contained natural-language experience",
      "confidence": 0.0
    }
  ]
}

Return {"experiences":[]} when the supplied trajectory source does not support a specific and reusable diagnostic lesson.
\end{lstlisting}
\prompttitle{Flat Experience Intervention}
\label{app:flat_online_prompt}

\begin{lstlisting}[style=prompt]
You are a doctor conducting a multi-turn clinical interaction. One unstructured experience distilled from a different historical case is provided as additional context. Follow the shared doctor-agent instructions and select exactly one action from the available tool schema at each turn.

Rules:
1. The retrieved experience may be irrelevant. Treat it only as transferable diagnostic guidance, never as evidence about the current patient.
2. Identify the clinically relevant lesson expressed by the experience and determine whether it is supported, contradicted, or not established by the current patient's observations.
3. Use the experience only when it is applicable to the current evidence. Semantic similarity alone is not sufficient to change the diagnostic process.
4. Base every action and diagnosis on the current patient's interaction history and patient-specific observations. Do not invent positive, negative, normal, or unavailable findings.
5. When sufficient patient-specific evidence has been acquired, submit one primary diagnosis at the specificity supported by the available evidence.

[CURRENT INTERACTION HISTORY]
[RETRIEVED FLAT EXPERIENCE]

Return exactly one tool call and no additional text.
\end{lstlisting}

\subsection{Evidence-Atom Normalization}
\label{app:normalization_prompt}

\begin{lstlisting}[style=prompt]
You are a clinical concept normalizer. Map each supplied clinical mention to a canonical evidence concept grounded in the Unified Medical Language System (UMLS), using the source span and its local context.

Normalization rules:
1. Preserve the clinical meaning of the supplied mention using only its source span and local context. Do not introduce findings or attributes that are not supported by the input.

2. Preserve assertion polarity, temporal status, numerical direction when applicable, and anatomical context. Negated, uncertain, or historical findings must not be represented as current positive evidence.

3. Treat only observed patient-specific findings as clinical evidence. A test request, order, plan, diagnostic hypothesis, retrieved knowledge statement, pending result, or unavailable result does not establish that a finding is present or absent in the current patient.

4. Map each mention to the most specific UMLS concept supported by both the mention and its context. Do not introduce a more specific finding, disease, anatomical site, or severity than the source text supports.

5. Return UNRESOLVED rather than forcing a weak, ambiguous, or conflicting UMLS match. When the concept is unresolved, return an empty CUI.

[CLINICAL MENTIONS WITH SOURCE SPANS AND LOCAL CONTEXT]
[CANDIDATE UMLS CONCEPTS WITH NAMES AND CUIS]

Return JSON only:
{
  "mentions": [
    {
      "concept": "...|UNRESOLVED",
      "cui": "...",
      "polarity": "present|absent|uncertain",
      "temporality": "current|historical|unknown",
      "numeric_direction": "high|low|normal|not_applicable",
      "anatomical_context": "...|none"
    }
  ]
}
\end{lstlisting}

\subsection{Counterfactual Validation Prompts}
\label{app:counterfactual_prompt}

For each candidate, both the original and intervened evidence contexts are evaluated twice. The first evaluation presents $(D_s,D_t)$ in this order, whereas the second reverses the order to $(D_t,D_s)$. For completion and reappraisal, a candidate is retained only when both diagnosis orders select $D_s$ before intervention and $D_t$ after intervention. For the blocking test, a candidate is retained only when both diagnosis orders select $D_t$ before intervention and $D_s$ after intervention. Candidates with order-dependent outputs or any required \texttt{UNRESOLVED} outcome are not retained.

\begin{lstlisting}[style=prompt]
You are a clinical diagnostic evidence evaluator. Determine which of the two candidate diagnoses is better supported by the supplied patient-specific evidence.

Use only the evidence provided below. Do not infer unobserved findings or use trajectory outcomes and reference labels.

CANDIDATE DIAGNOSES:
1. [DIAGNOSIS 1]
2. [DIAGNOSIS 2]

EVIDENCE CONTEXT:
[ORIGINAL PATIENT EVIDENCE]

Select the diagnosis better supported by the complete supplied evidence context. Return UNRESOLVED if the evidence does not distinguish the two diagnoses.

Return only:
[DIAGNOSIS 1 | DIAGNOSIS 2 | UNRESOLVED]
\end{lstlisting}

\prompttitle{Completion Test}
\label{app:completion_test_prompt}

\begin{lstlisting}[style=prompt]
You are a clinical diagnostic evidence evaluator. Determine which of the two candidate diagnoses is better supported by the supplied patient-specific evidence.

Use only the evidence provided below. Do not infer unobserved findings or use trajectory outcomes and reference labels.

CANDIDATE DIAGNOSES:
1. [DIAGNOSIS 1]
2. [DIAGNOSIS 2]

EVIDENCE CONTEXT:
[ORIGINAL PATIENT EVIDENCE WITH ONE EVIDENCE ATOM ADDED]

EVIDENCE ADDED:
[CANDIDATE EVIDENCE ATOM]

EVIDENCE ADDED is the only change from the original evidence context. Evaluate its diagnostic significance together with all other patient evidence. Do not assume that it supports either diagnosis.

Select the diagnosis better supported by the complete supplied evidence context. Return UNRESOLVED if the evidence does not distinguish the two diagnoses.

Return only:
[DIAGNOSIS 1 | DIAGNOSIS 2 | UNRESOLVED]
\end{lstlisting}
\prompttitle{Reappraisal Test}
\label{app:reappraisal_test_prompt}

\begin{lstlisting}[style=prompt]
You are a clinical diagnostic evidence evaluator. Determine which of the two candidate diagnoses is better supported by the supplied patient-specific evidence.

Use only the evidence provided below. Do not add, remove, or alter any patient finding. Do not use trajectory outcomes or reference labels.

CANDIDATE DIAGNOSES:
1. [DIAGNOSIS 1]
2. [DIAGNOSIS 2]

EVIDENCE CONTEXT:
[THE SAME COMPLETE OBSERVED PATIENT EVIDENCE]

EVIDENCE TO REAPPRAISE:
[CANDIDATE EVIDENCE ATOM ALREADY PRESENT IN THE CONTEXT]

Pay particular attention to EVIDENCE TO REAPPRAISE. Reconsider its diagnostic significance when interpreted together with the complete supplied evidence context. Do not assume that it supports either diagnosis.

Select the diagnosis better supported by the complete supplied evidence context. Return UNRESOLVED if the evidence does not distinguish the two diagnoses.

Return only:
[DIAGNOSIS 1 | DIAGNOSIS 2 | UNRESOLVED]
\end{lstlisting}
\prompttitle{Blocking Test}
\label{app:blocking_test_prompt}

\begin{lstlisting}[style=prompt]
You are a clinical diagnostic evidence evaluator. Determine which of the two candidate diagnoses is better supported by the supplied patient-specific evidence.

Use only the evidence provided below. Do not infer unobserved findings or use trajectory outcomes and reference labels.

CANDIDATE DIAGNOSES:
1. [DIAGNOSIS 1]
2. [DIAGNOSIS 2]

EVIDENCE CONTEXT:
[ORIGINAL PATIENT EVIDENCE WITH ONE EVIDENCE ATOM ADDED]

EVIDENCE ADDED:
[CANDIDATE EVIDENCE ATOM]

EVIDENCE ADDED is the only change from the original evidence context. Evaluate its diagnostic significance together with all other patient evidence. Do not assume that it supports either diagnosis.

Select the diagnosis better supported by the complete supplied evidence context. Return UNRESOLVED if the evidence does not distinguish the two diagnoses.

Return only:
[DIAGNOSIS 1 | DIAGNOSIS 2 | UNRESOLVED]
\end{lstlisting}

\subsection{Evidence-State Intervention Prompts}
\label{app:evidence-state_prompt}

\prompttitle{Missing-Evidence State}
\label{app:missing_prompt}

\begin{lstlisting}[style=prompt]
The current evidence state does not yet contain an observation corresponding to a validated decision-critical evidence item. Acquiring the corresponding patient-specific observation may help resolve the diagnostic boundary. 

Evidence category to acquire:
[EVIDENCE CATEGORY]

Recommended acquisition action:
[ACQUISITION ACTION]

Acquire the missing evidence using the recommended action. Do not assume whether the resulting observation will be positive, negative, normal, or abnormal. Treat only the patient-specific observation returned by the diagnostic environment as evidence, and update the diagnosis using the complete interaction history.
\end{lstlisting}

\prompttitle{Ignored-Evidence State}
\label{app:ignored_prompt}

\begin{lstlisting}[style=prompt]
The current evidence state already contains validated decision-critical evidence that may not have been sufficiently incorporated into the proposed diagnosis.

Observed decision-critical evidence for reappraisal: [OBSERVED PATIENT-SPECIFIC EVIDENCE]
Retrieved diagnosis: [DIAGNOSIS]

Reassess the proposed diagnosis using the complete interaction history, with particular attention to the diagnostic implications of the highlighted evidence. Determine whether this evidence weakens the proposed diagnosis and whether the retrieved diagnosis provides a better explanation of the available patient-specific evidence.

This guidance supports evidence reappraisal rather than prescribing a diagnosis. Do not introduce unobserved findings or treat retrieved general knowledge as evidence about the current patient. Retain or revise the diagnosis according to the complete patient-specific evidence.

\end{lstlisting}

\subsection{Evaluation and Audit Prompts}
\label{app:evaluation_prompts}

\prompttitle{Diagnostic Evaluation Rubric}
\label{app:judge_prompt}

\begin{lstlisting}[style=prompt]
You are a medical expert evaluating the diagnostic accuracy of a single-answer clinical diagnosis task.

Determine whether the model diagnosis is clinically consistent with the reference diagnosis. Judge whether they represent the same diagnostic boundary rather than whether they are merely related.

Answer YES when the model diagnosis:
1. denotes the same diagnosis as the reference, including a true synonym, abbreviation, spelling variant, eponym, or clinically equivalent formulation; or
2. provides a more specific diagnosis that is fully compatible with the reference and does not contradict any boundary-defining attribute.

Answer NO when the model diagnosis:
1. denotes only a broader parent category or generic disease class;
2. denotes only a symptom, sign, manifestation, complication, or differential diagnosis;
3. denotes a different disease, etiology, mechanism, anatomical site, subtype, or stage;
4. omits a boundary-defining specification required by the reference diagnosis; or
5. is merely associated with, compatible with, or clinically related to the reference diagnosis.

Reference diagnosis:
[REFERENCE]

Model diagnosis:
[PREDICTION]

Are the two diagnoses clinically consistent under the same diagnostic boundary?

Answer with exactly one word: YES or NO.
\end{lstlisting}
\paragraph{Evidence-Sufficiency Audit Prompts.}
\label{app:audit_prompts}

\prompttitle{Blind Evidence Parser}
\begin{lstlisting}[style=prompt]
Extract the patient-specific clinical evidence observed before the final diagnostic action. Perform this extraction without access to the reference diagnosis.

Rules:
1. Include only findings explicitly returned by the patient or diagnostic environment, including available history, examination findings, laboratory results, EHR information, and imaging findings.
2. Do not treat a test request, order, plan, diagnostic hypothesis, model reasoning, retrieved general knowledge, pending result, or unavailable result as observed patient evidence.
3. Preserve the polarity, uncertainty, and temporal status of each observation. Negated, uncertain, or historical findings must not be represented as current positive evidence.
4. Do not infer an unobserved finding from the doctor agent's diagnosis, differential diagnosis, requested examination, or general medical knowledge.
5. Merge repeated mentions of the same clinical finding while preserving clinically meaningful differences in polarity, temporality, numerical direction, and anatomical context.

[PRE-FINAL ACTION--OBSERVATION HISTORY, EXCLUDING THE FINAL DIAGNOSTIC ACTION]
Return JSON only:
{
  "observed_evidence": [
    {
      "evidence": "...",
      "polarity": "present|absent|uncertain",
      "temporality": "current|historical|unknown",
      "numeric_direction": "high|low|normal|not_applicable",
      "anatomical_context": "...|none"
    }
  ]
}
\end{lstlisting}

\prompttitle{Evidence-Sufficiency Adjudicator}
\begin{lstlisting}[style=prompt]
Given the blind evidence parse, complete pre-final action--observation
history, vanilla agent diagnosis, and reference diagnosis, determine whether the patient-specific evidence observed before the final diagnostic action was sufficient to favor the reference diagnosis.

Use only patient-specific evidence observed in the trajectory. General medical knowledge may support interpretation but must not be treated as evidence about the current patient. Do not infer unavailable findings or assume that missing evidence is positive or negative.

Determine:
1. the key observed evidence supporting the reference diagnosis;
2. important discriminative evidence that remained unobserved;
3. observed evidence supporting the vanilla agent diagnosis or a competing diagnosis; and
4. the final evidence-sufficiency state.

Classification rules:
- INCOMPLETE: important discriminative evidence needed to favor the reference diagnosis remained unobserved, so additional patient-specific evidence would have been required.
- SUFFICIENT: the observed evidence was sufficient to favor the reference diagnosis over the vanilla agent diagnosis, but this evidence was not adequately incorporated into the final diagnosis.
- CONFLICTING: the observed evidence materially supported competing diagnoses and did not reliably resolve the diagnostic boundary.

Assign exactly one state. Use SUFFICIENT only when the reference diagnosis can be favored from the already observed evidence without requiring additional patient-specific evidence. Use CONFLICTING rather than SUFFICIENT when the available evidence does not reliably distinguish the reference diagnosis from a plausible competing diagnosis.

[BLIND EVIDENCE PARSE]
[COMPLETE PRE-FINAL ACTION--OBSERVATION HISTORY]
[VANILLA AGENT DIAGNOSIS]
[REFERENCE DIAGNOSIS]

Return JSON only:
{
  "key_supporting_evidence": ["..."],
  "missing_discriminative_evidence": ["..."],
  "competing_or_conflicting_evidence": ["..."],
  "final_evidence_state": "INCOMPLETE|SUFFICIENT|CONFLICTING"
}
\end{lstlisting}
% =========================================================
% D. Ethics Statements
% =========================================================
\section{Data Ethics Statement}
\label{app:data_ethics}

\paragraph{Data governance.}
MIMIC-IV, MIMIC-IV-Note, and MIMIC-CXR are de-identified,
credentialed-access datasets governed by the applicable PhysioNet data
use agreements. All data were accessed and used in accordance with the
corresponding credentialing requirements and terms of use.

The AgentClinic MedQA subset is derived from previously released
benchmark data and is used solely for research evaluation.
\end{document}